\documentclass[runningheads]{llncs}
\usepackage{graphicx}
\usepackage{amssymb}
\usepackage{amsmath}
\usepackage{hyperref}
\usepackage{comment}
\usepackage{listings}
\usepackage{xcolor}
\usepackage[most]{tcolorbox}

\newcommand{\sheeba}[1] {\textbf{\textcolor{blue}{~#1}}}
\newcommand{\vamsi}[1] {\textbf{\textcolor{red}{~#1}}}
\newcommand{\bkr}[1] {\textbf{\textcolor{brown}{~#1}}}

\begin{document}
\title{From human experts to machines: An LLM supported approach to ontology and knowledge graph construction\thanks{Supported by the German Centre for Integrative Biodiversity Research (iDiv) Halle-Jena-Leipzig, funded by the German Research Foundation (FZT 118, 202548816) and the Carl Zeiss Foundation project  `A Virtual Werkstatt for Digitization in the Sciences(K3)' within the scope of the program line `Breakthroughs: Exploring Intelligent Systems for Digitization - explore the basics, use applications'}}
\titlerunning{An LLM supported approach to ontology and knowledge graph construction}
\author{Vamsi Krishna Kommineni\inst{1,2,3}\orcidID{0000-0001-6168-3085} \and
Birgitta König-Ries\inst{1,2,4}\orcidID{0000-0002-2382-9722} \and
Sheeba Samuel\inst{1,4}\orcidID{0000-0002-7981-8504}}
\authorrunning{VK. Kommineni et al.}
\institute{Heinz Nixdorf Chair for Distributed Information Systems, Friedrich Schiller University Jena, Jena, Germany \and German Centre for Integrative Biodiversity Research (iDiv) Halle-Jena-Leipzig, Leipzig, Germany \and Max Planck Institute for Biogeochemistry, Jena, Germany \and
Michael Stifel Center Jena, Jena,  Germany \\
\email{(vamsi.krishna.kommineni, birgitta.koenig-ries, sheeba.samuel)@uni-jena.de }}
\maketitle              % typeset the header of the contribution
\begin{abstract}
The conventional process of building Ontologies and Knowledge Graphs (KGs) heavily relies on human domain experts to define entities and relationship types, establish hierarchies, maintain relevance to the domain, fill the ABox (or populate with instances), and ensure data quality (including amongst others accuracy and completeness). On the other hand, Large Language Models (LLMs) have recently gained popularity for their ability to understand and generate human-like natural language, offering promising ways to automate aspects of this process. This work explores the (semi-)automatic construction of KGs facilitated by open-source LLMs. Our pipeline involves formulating competency questions (CQs), developing an ontology (TBox) based on these CQs, constructing KGs using the developed ontology, and evaluating the resultant KG with minimal to no involvement of human experts. We showcase the feasibility of our semi-automated pipeline by creating a KG on deep learning methodologies by exploiting scholarly publications. To evaluate the answers generated via Retrieval-Augmented-Generation (RAG) as well as the KG concepts automatically extracted using LLMs, we design a judge LLM, which rates the generated content based on ground truth. Our findings suggest that employing LLMs could potentially reduce the human effort involved in the construction of KGs, although a human-in-the-loop approach is recommended to evaluate automatically generated KGs.
\end{abstract}
\section{Introduction}

 In information organization and representation, ontologies stand as foundational frameworks for describing and structuring domain knowledge. These structured representations not only represent the entities and relationships within a domain but also serve as the foundation for constructing comprehensive knowledge graphs (KGs) \cite{hogan2021knowledge}. KGs, in turn, offer a powerful mechanism for interlinking diverse pieces of information and facilitating sophisticated data analytics and reasoning. The domain knowledge encapsulated within ontologies constitutes a valuable asset for various knowledge-intensive applications. This comes at a price, though: Ontology and knowledge engineering represents a collaborative and interdisciplinary effort, demanding the time and expertise of multiple stakeholders \cite{funk2023towards,pan2023large}. The higher the expressiveness in the ontology language, the more intricate design choices need to be made throughout the construction process with additional developmental challenges including accuracy, scalability, and depth of knowledge captured.
The conventional approach to constructing KGs typically involves gathering domain requirements through CQs (also proposed as the requirement specification in ontology development) collected from domain experts, collaborating with computer scientists and domain specialists to develop an ontology, transforming unstructured data into structured formats, and finally, populating the ontology to create the KG \cite{haussmann2019foodkg}. In this context, a pressing question emerges: How can we strike a balance between the resource-intensive nature of ontology construction and the imperative of leveraging domain knowledge effectively? 

This research paper attempts to address this question by leveraging Large Language Models (LLMs) in Knowledge Graph Engineering, particularly focusing on minimizing the time and human effort involved in these processes. In recent years, the emergence of these models has revolutionized the landscape of natural language processing (NLP) and knowledge representation.  Equipped with massive pre-trained parameters and advanced neural architectures, LLMs exhibit remarkable capabilities in understanding and generating human-like text across a spectrum of languages and domains \cite{brown2020language,chowdhery2023palm}. As a result, they have gained immense popularity and adoption across diverse fields ranging from information retrieval to language translation.

Hence, we explore the (semi-)automatic construction of a KG, starting from collecting competency questions (CQs) to creating an ontology and to filling the data in the ontology, facilitated by the integration of LLMs. 
%The conventional approach to constructing KGs typically involves gathering domain requirements through CQs (also proposed as the requirement specification in ontology development) collected from domain experts, collaborating with computer scientists and domain specialists to develop an ontology, transforming unstructured data into structured formats, and finally, populating the ontology to create the KG \cite{haussmann2019foodkg}. In our study, we adhere to the fundamental steps of ontology and KG construction; however, we streamline the process by automating the key tasks using LLMs.
To test the feasibility of our approach, we apply this methodology to %the domain of 
creating an ontology and KGs about deep learning (DL) methodologies extracted from scholarly publications in the biodiversity domain.
%as we outlined in \cite{samuel2020machine}. 

The choice of this example has been motivated by the increasing usage of DL in research. 
%The utilization of machine learning (ML) and DL techniques has seen a significant rise across various domains, including but not limited to life-sciences.
%Ensuring the reproducibility of outcomes generated by DL pipelines holds paramount importance to validate and trust the results obtained. Therefore, 
Documenting the provenance of DL results is indispensable to facilitate the reproducibility of these studies, a prerequisite to trust and validation of results. To accomplish this task effectively,  information, such as models, architectures, hyperparameters, and other key details needs to be captured and stored in a structured representation.
\section{Related work}
\label{sec:relatedwork}
LLMs have revolutionized knowledge engineering and NLP, showcasing human-level performance across diverse linguistic tasks \cite{brown2020language,chowdhery2023palm}.  
With the increasing robustness of LLMs, their potential as a knowledge source in various applications such as KG completion, ontology refinement, and question answering has become evident \cite{liu2020concept,petroni2019language,yao2019kg}. Research has rapidly expanded to explore the application of LLMs, with recent papers providing surveys on the use of LLMs in KG engineering along with associated challenges \cite{meyer2023llm,pan2023large}.
Meyer et al. \cite{meyer2023llm} presents a list of application areas for LLM-assisted KG engineering, including creating or enriching KG schemas/ontologies, populating KGs, etc.
In their position paper, Pan et al. \cite{pan2023large} present opportunities, visions, research topics, and challenges for LLMs for KGs and KGs for LLMs.

Recent studies have introduced methods for ontology creation \cite{cohen2023crawling,funk2023towards}, augmentation \cite{zaitoun2023can}, completion \cite{toro2023dynamic}, and learning \cite{babaei2023llms4ol} using LLMs.
Cohen et al. \cite{cohen2023crawling} present a crawling approach to extract a KG from an LLM using ‘subject-relation-object’ statement formats. Funk et al. \cite{funk2023towards} focus on constructing a concept hierarchy for a given domain starting from a seed concept by querying LLMs. However, they consider only subconcept/is-a relation, but no other relations. In this work, we focus on reusing existing ontologies with all their concepts and relations.
While there is limited research on the utilization of LLMs for completing KGs or ontologies \cite{veseli2023evaluating}, this trend appears to be changing rapidly.
While LLMs offer new possibilities for ontology learning and development, they do not alter the fundamental need for expert collaboration and establishing consensus within a community. However, they may enhance the productivity of ontologists by simplifying ontology development and integrating into semi-automatic toolchains \cite{neuhaus2023ontologies}.
In our approach, we have reused the existing ontology in the domain and involved humans in the loop to validate and evaluate the LLM-generated content.

%Most publications use GPT 3.5 and 4 to exhibit their work, which depends on LLMs 
Most of the publications discussed above \cite{funk2023towards,meyer2023llm,neuhaus2023ontologies,toro2023dynamic,zaitoun2023can} are based on openAI's GPT 3.5\footnote{\url{https://platform.openai.com/docs/models/gpt-3-5}} or 4\footnote{\url{https://platform.openai.com/docs/models/gpt-4}}. However, the usage cost of OpenAI API models can escalate quickly if deployed for large-scale applications. In contrast, open-source LLMs offer transparency, model control, usage flexibility, and cost-effectiveness. 
%We have implemented our complete pipeline using open-source LLMs to leverage the advantage of open-source LLMs. 
As far as our knowledge extends, this represents the first approach to introduce a comprehensive (semi-)automated pipeline for constructing ontology and KGs by using open-source LLMs.
%, with the aim of extracting DL techniques from scholarly publications for reproducibility.
\begin{comment}
    BKR: this is a very specific description. Is the restrictuion to "the aim of extracting..." necessary? or could you state more broadly "first ... ontology and KGs from scholarly publications using open-source LLMs"
    \sheeba{Change to the broad description.}
\end{comment}
%\sheeba{Add 1-2 sentence on open source LLMs and OpenAi and their limitations. Current literature shows examples with chatgpt examples and openai. why we need to use open source llms to scale our work.}
\begin{comment}    
\begin{itemize}
    \item General/Traditional pipeline in Ontology and KG creation
    \item LLM use in Semantic web 
    \item LLM in Ontology \cite{babaei2023llms4ol,zaitoun2023can,toro2023dynamic}
    \item LLM in KG Engineering \cite{meyer2023llm}, challenges \cite{pan2023large}
\end{itemize}
\end{comment}
\section{Methods}
\label{sec:methods}
\begin{figure}
\centering
\includegraphics[width=\textwidth]{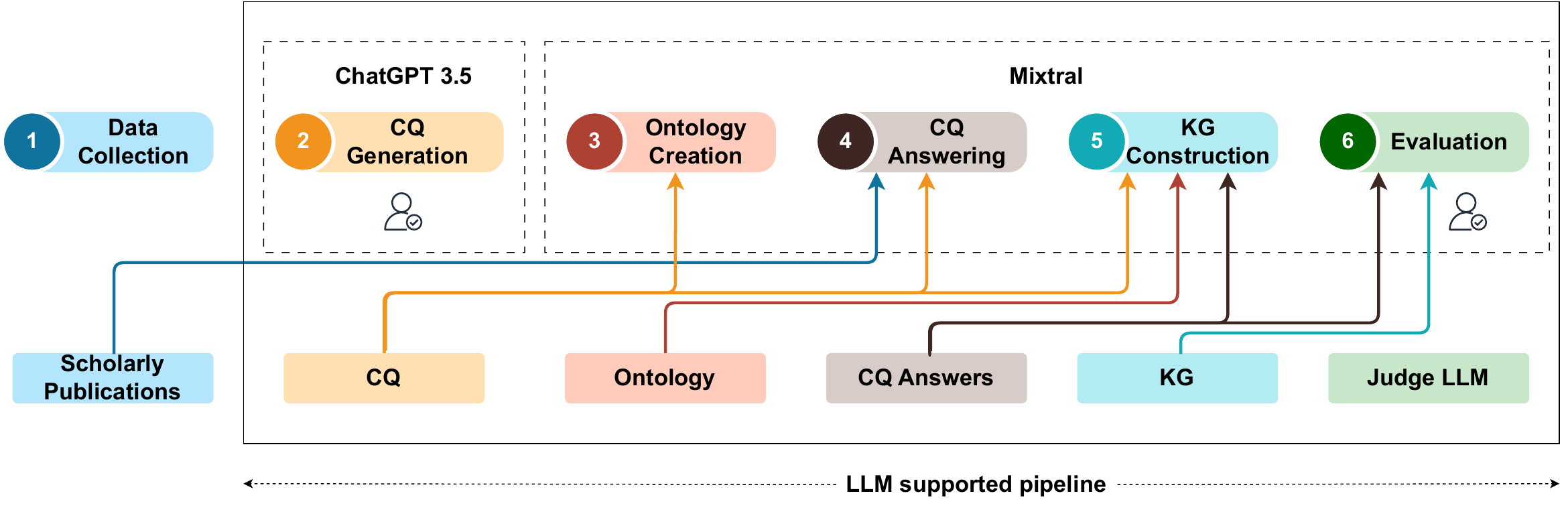}
\caption{The (semi-)automatic approach for constructing KGs. Human-in-the-loop is involved in the second and sixth stages of the LLM supported pipeline.} %\label{kgpipeline}
\label{fig:methods}
\end{figure}

This section presents our (semi-)automatic pipeline for constructing KGs with six main components: Data Collection, CQ generation, Ontology creation, CQ answering, KG construction, and Evaluation (Figure \ref{fig:methods}). % shows the framework of our proposed pipeline. 
The prompts, code, results, and data used for the evaluation are publicly available at GitHub: \url{https://github.com/fusion-jena/automatic-KG-creation-with-LLM}. The computational infrastructure to create and execute the pipeline is provided by Friedrich Schiller University Jena: Draco cluster, computer node with GPU accelerator, 1x NVIDIA A100, 80GB.
%\url{https://github.com/fusion-jena/dlkgllm}. 

%\vamsi{github link is not working, do you have access to fusion github? if yes create the link and provide me access , I will add all the necessary files and data}
%\sheeba{I have not created the repo. Is 'dlkgllm' as repository name fine or something more elaborative. I can create the link and will add you.} 
%\vamsi{maybe "Automatic KG-creation with LLM", will also be useful for search indexing via google. \sheeba{Sounds good}}

%Each component is briefly described below:

%\sheeba{Here is the link to the figure creation. I used draw.io: \url{https://drive.google.com/file/d/1j7M1pilChcMZ-zn8xwZHw_sHnobp_HYM/view?usp=sharing} If you cannot access, let me know what changes need to be made, I can do it.} \vamsi{Accessed it, only added one more connection from CQ to KG Construction} 

 \textbf{Data Collection:} We reused a dataset generated in our prior research \cite{ahmed2023reproducible}, where we conducted a systematic literature review to identify publications employing DL methods in biodiversity research based on keywords suggested by biodiversity experts \cite{abdelmageed2022biodivnere}. Additionally, two domain experts curated a dataset of 61 publications and manually extracted reproducibility-related variables inspired by the current literature\footnote{The dataset with reproducibility-related variables is available at \url{https://github.com/fusion-jena/Reproduce-DLmethods-Biodiv}}. We used the first five of these 61 publications in this work to test our (semi-) automated pipeline. \newline
 \textbf{CQ generation:} To generate the CQs, we prompted ChatGPT-3.5 to get abstract-level questions to describe the provenance of the results of DL pipelines. Two human domain experts evaluated the CQs generated by the ChatGPT-3.5 web interface to enhance existing questions and add new ones. 
    %Through this process, 40 CQs were created and subsequently used for ontology creation. \sheeba{This is a part of the result, so goes in the result section}
   \begin{comment}
   \bkr{
       BKR: I don't understand the roles of chatgpt and the open source models. 
       Also, I am not sure whether the exact parameters need to be reported in the paper or whether it would be sufficient to refer to the github for this
   }
   \end{comment}

    After the CQ generation step, we conducted experiments using three open-source models: Llama 2-70B\footnote{\url{https://huggingface.co/meta-llama/Llama-2-70b-chat-hf}}, `Mixtral 8x7B'\footnote{\url{https://mistral.ai/news/mixtral-of-experts/}}, and Falcon-40B\footnote{\url{https://huggingface.co/tiiuae/falcon-40b}}. Upon manual comparison of the content quality generated by these models, it became apparent that the outputs of Mixtral 8x7B were superior comparatively and provided the expected results. Hence, starting from the ontology creation component, we employed `Mixtral 8x7B', a sparse mixture-of-experts network, which was inferred at a zero-shot setting.    
    The hyperparameters of zero-shot Mixtral 8x7B include a temperature of ($10^{-5}$) and a maximum output token limit of 25,000 tokens. For the RAG pipeline, the chunk size for text splitting is set at 2,500 tokens, with a chunk overlap of 100 tokens.
    %For any specific implementation details and further hyperparameters, check our \href{https://github.com/fusion-jena/automatic-KG-creation-with-LLM}{publicly available code}. \sheeba{Commenting out to save space, since it is already provided} 
    %The computational infrastructure to create and execute the pipeline is provided by Friedrich Schiller University Jena: Draco cluster, computer node with GPU accelerator, 1x NVIDIA A100, 80GB.
%\sheeba{Add more information on the settings of the open source LLMs, e.g., temperature, tokens}    
%\sheeba{Add information on the hardware infrastructure used: Draco cluster, computer node with GPU accelerator, 1x NVIDIA A100, 80GB}
\begin{comment}    
\begin{tcolorbox}[enhanced,attach boxed title to top center={yshift=-3mm,yshifttext=-1mm},
  colback=blue!5!white,colframe=blue!75!black,colbacktitle=blue!80!black,
  title=Competency questions generation prompt,fonttitle=\bfseries,
  boxed title style={size=small,colframe=blue!50!black} ]
  Write the competency questions based on the abstract level concepts for describing the provenance of DL pipeline
\end{tcolorbox}
\end{comment}   
\newline
 \textbf{Ontology creation:}
\begin{comment}
    \begin{tcolorbox}[enhanced,attach boxed title to top center={yshift=-3mm,yshifttext=-1mm},
  colback=blue!5!white,colframe=blue!75!black,colbacktitle=blue!80!black,
  title=concepts and relationships extraction prompt, fonttitle=\bfseries,
  boxed title style={size=small,colframe=blue!50!black} ]
           INSTRUCTIONS: \newline
            Analyze the following competency questions and identify all the concepts and relationships between concepts mentioned in the text.
            These concepts will be used to build ontology for describing the provenance of DL Pipeline.
            If you don't know the answer, just say that you don't know, don't try to make up an answer.
            Don't provide anything other than the results in comma separated list.
            Do not reply using a complete sentence, and only give the answer in the following format: \newline
            Below are the examples and follow the same format to answer all the questions: \newline
            CQ1: What hyperparameters are used in the model? \newline
            CQ2: What data formats are used in the deep learning pipeline? \newline
            CQ3: What are the sources of input data for the deep learning pipeline? \newline
            Concepts: Hyperparameter, Model, Data, DataFormat, DeepLearningPipeline, Source, InputData \newline
            Relations: hasHyperparameter, hasModel, hasData, hasDataFormat, hasSource, hasInputData \newline
        QUERY: \newline
        query : {CQs}
\end{tcolorbox}

\end{comment}    
    A two-step strategy was implemented to create the ontology from the LLM-generated human-verified CQs. In the first step, we aimed to extract all concepts and their relationships from the CQs. To achieve this, we included a set of instructions and an example CQ with expected output in the prompt. In the second step, we constructed an ontology for describing information on DL pipelines. This was achieved by providing an in-context example containing a basic ontology structure, utilizing the PROV-O ontology \cite{lebo2013prov} as a foundational ontology for reuse, and incorporating the concepts and relationships extracted from the CQs.\\
\begin{comment}    
\begin{tcolorbox}[enhanced,attach boxed title to top center={yshift=-3mm,yshifttext=-1mm},
  colback=blue!5!white,colframe=blue!75!black,colbacktitle=blue!80!black,
  title=Ontology creation prompt,fonttitle=\bfseries,
  boxed title style={size=small,colframe=blue!50!black} ]
  INSTRUCTIONS: \newline
  Use the concepts and relations (properties) and build an ontology in RDF format for describing the provenance of DL pipeline. If you don't know the answer, just say that you don't know, don't try to make up an answer. Don't provide anything other than the ontology in RDF format.        Use the IRI for the base ontology: https://w3id.org/dlprov/. Use the ontology classes and relations as the base ontology. Below are the examples and follow the same format for all the questions:
  
        Concepts: Hyperparameter, Model
        
        Relations: hasHyperparameter, hasModel \newline
        QUERY: \newline
        concepts : {concepts} \newline
        relations: {relations} \newline
\end{tcolorbox}
\end{comment}    
\textbf{CQ Answering:} This component is the central pillar of the whole pipeline, particularly for KG construction. With this component, we retrieved answers for all the CQs using the RAG approach from the first five selected biodiversity scholarly publications from our dataset that employed DL methods. We then applied basic text processing techniques to refine the generated answers, eliminating redundant and repetitive content where applicable.
\begin{comment}        
\begin{tcolorbox}[enhanced,attach boxed title to top center={yshift=-3mm,yshifttext=-1mm},
  colback=blue!5!white,colframe=blue!75!black,colbacktitle=blue!80!black,
  title=Competency Question Answering prompt,fonttitle=\bfseries,
  boxed title style={size=small,colframe=blue!50!black} ]
INSTRUCTIONS:

Use the provided pieces of context to answer the query. If you don't know the answer, just say that you don't know, don't try to make up an answer.

Query: "{query}"

\end{tcolorbox}
\end{comment}   
\newline
\textbf{KG construction:}
To construct the KG, processed CQ answers along with their respective questions and the LLM-generated ontology were given as input to the LLM. With the prompt, we instructed the LLM to extract key entities, relationships, and concepts from the answers and map them onto the ontology to generate the KG.\\
\begin{comment}
    \begin{tcolorbox}[enhanced,attach boxed title to top center={yshift=-3mm,yshifttext=-1mm},
  colback=blue!5!white,colframe=blue!75!black,colbacktitle=blue!80!black,
  title=Population of KG prompt,fonttitle=\bfseries,
  boxed title style={size=small,colframe=blue!50!black} ]  
  Your task is to create a knowledge graph based on the provided competency questions, answers, and ontology (is in OWL format). Only by using the provided content, create a knowledge graph in RDF format, don't try to make up an answer. Below are the competency questions and answers: 

  $<$CQs$>$

   $<$CQ Answers$>$

Below is the ontology:

  $<$Ontology in OWL format$>$
\end{tcolorbox}  

\end{comment}    
\textbf{Evaluation:}
Two key outputs produced by the LLM were evaluated in this step: the generated CQ answers and the KG concepts that were automatically extracted from these answers.
To evaluate these outputs, we employed the LLM as a judge (instructed using a prompt) to assign scores to the generated content based on human evaluator-generated ground truth.  Due to the time and human effort required for manual annotation, a human expert annotated only five scholarly articles with ground truth text for the CQs and assigned single words (Right, Wrong, and Partial) to evaluate the RAG-generated CQ answers. Consequently, these five publications are used in our complete pipeline to test the feasibility of our approach. 
\newline
To assess the generated CQ answers, we asked the LLM to score the alignment between the ground truth and the generated answer on a scale from 0 to 10, where 0 represents no alignment and 10 represents maximum alignment. Then we classified scores greater than or equal to six as Right, scores less than three as Wrong, and the remaining scores as Partial.
The automatically extracted KG concepts were also evaluated with the judge LLM, which checked whether the KG individual concepts appeared in the respective generated CQ answers. We used the ontology concepts to establish links between the KG individual concept and the corresponding generated CQ answers.

\section{Results}
\label{sec:results}
%In this section, we present the results from each stage of our pipeline.
\begin{comment}
    who is the we in we formulated? sounds as though that was done by you personally not your pipeline. Maybe rather: 'The CQ generation step of the pipeline resulted in 40 CQs.' Similar in the following steps.
\end{comment}
The CQ generation step of the pipeline resulted in 40 CQs. These CQs investigate every step of the DL pipeline, including raw data sources, preprocessing techniques, model architectures, hyperparameter settings, software and hardware choices, post-processing steps, security measures in handling sensitive data, and data biases alongside ethical considerations.
\vspace{-0.5cm}
\begin{figure}
\centering
\includegraphics[width=11cm]{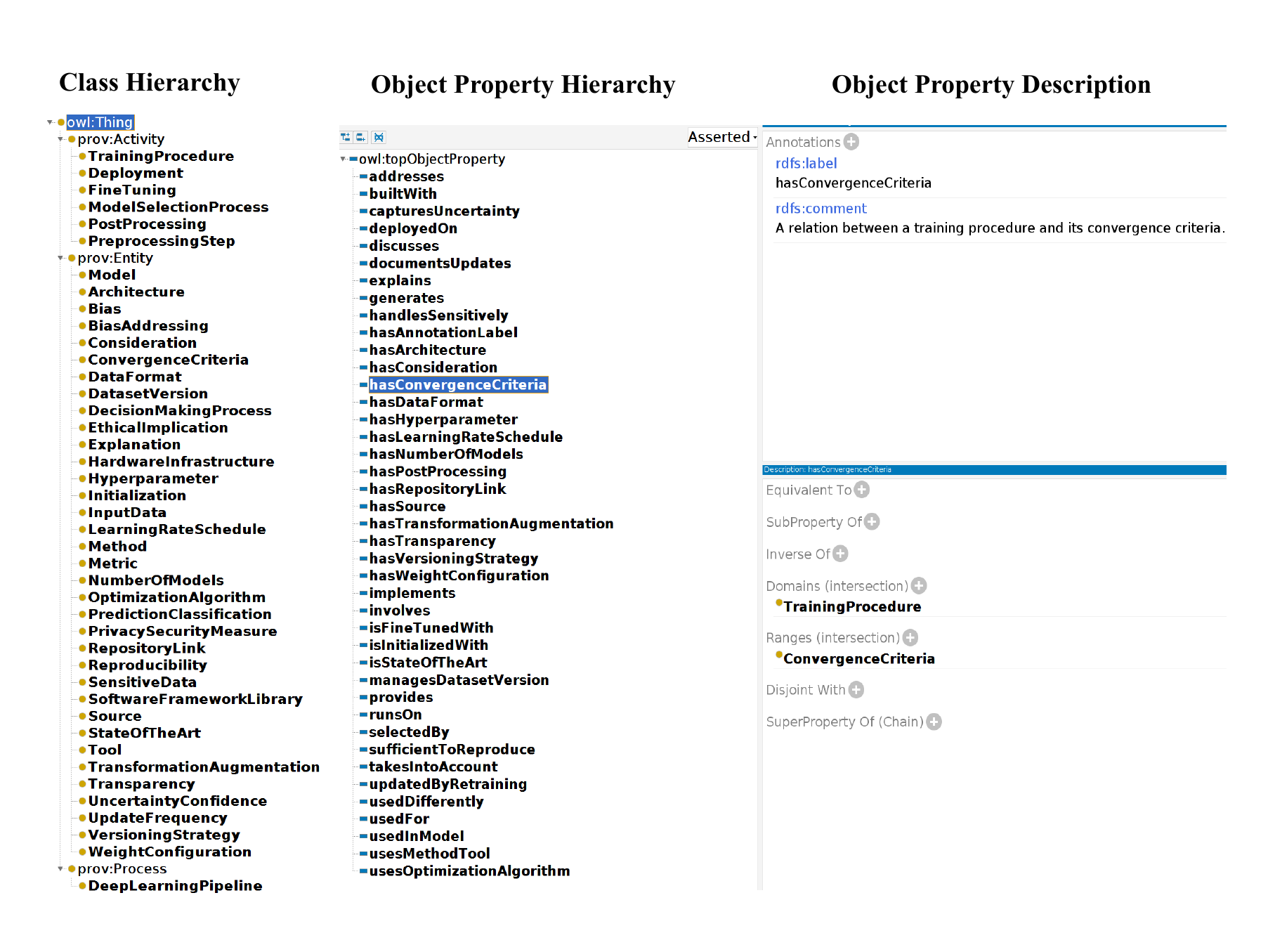}
\caption{An excerpt of the DLProv ontology generated by the LLM in our pipeline.} %\label{kgpipeline}
\label{fig:onto}
\end{figure}

Based on these CQs, the Ontology generation step of the pipeline developed the DLProv Ontology reusing the PROV-O ontology \cite{lebo2013prov}.
It consists of 45 classes and 41 relationships with 365 axioms. See Figure \ref{fig:onto} for an excerpt.
However, concepts were added as subclasses to three PROV-O classes only, and no object properties were reused.
In another experiment where we provided the MLSchema ontology\footnote{\url{http://www.w3.org/2016/10/mls/}} for reuse, only one class was reused and no object properties.

The CQ answering step of the pipeline generated answers for each CQ for the first five publications from our dataset. There were 42 disagreements between the human annotator and the LLM Judge out of 200 evaluated CQ answers during the evaluation of these RAG-generated answers. 

Using the CQs, the answers from the CQs generated from the scholarly publications and the ontology, a KG was constructed from the method information of DL pipelines extracted from five scholarly publications using our pipeline. Listing \ref{lst:KG-excerpt} shows the excerpt of the KG describing the provenance of the results from the DL methods from the publication \cite{klein2015deep}.
\begin{lstlisting}[caption={An excerpt of the KG generated by our (semi-) automated pipeline},label={lst:KG-excerpt}]
dlprov:DeepLearningPipeline_1 rdf:type dlprov:DeepLearningPipeline ;
    dlprov:hasDataFormat dlprov:DataFormat_1 ;
    dlprov:hasDataFormat dlprov:DataFormat_2 ;
dlprov:DataFormat_1 rdf:type dlprov:DataFormat ;
    rdfs:label 'Audio Spectrogram'.

dlprov:DataFormat_2 rdf:type dlprov:DataFormat ;
    rdfs:label 'Image data'.    
\end{lstlisting}
We have created multiple KGs in combination with two different prompts and two different RAG-generated CQ answers for five selected publications (Table \ref{tab:KGevalresult}). Our best KGs have successfully correlated 142 KG individuals with the CQ answers out of a total of 203 KG individuals across five KGs, each corresponding to a selected scholarly publication.
\begin{comment}
    \begin{table}[htbp]
\centering
\begin{tabular}{| p{4.5cm}| p{1.5cm} | p{1.5cm} | p{1.5cm} | p{1.5cm} |} 
 \hline
 Publication / Combination of prompt and CQ answers & Prompt v1 and CQ answer v1 & Prompt v1 and CQ answer v2 & Prompt v2 and CQ answer v1 & Prompt v2 and CQ answer v2 \\ 
 \hline\hline
 Klein et al. 2015 \cite{klein2015deep} & 24.32 \vamsi{56.76} & x & x & 9.42 \vamsi{8.77} \\ 
 \hline
 Khalighifar et al. 2021 \cite{Khalighifar2021-ad}  & 85.71 & x & x & 68.06 \vamsi{70.83} \\
 \hline
 Choe et al. 2021 \cite{Choe2021-ar} & 73.21 & 64.41 \vamsi{59.32} & 59.26 & 64.29 \vamsi{58.93} \\
 \hline
 Mahmood et al. 2016 \cite{Mahmood2016-nx} & 91.53 & 81.48 \vamsi{85.19} & 82.76 & 77.55 \vamsi{81.63} \\
 \hline
 Younis et al. 2020 \cite{Younis2020-ce} & 66.67 & x & 67.51 & 61.67 \vamsi{56.67} \\ [1ex] 
 \hline
\end{tabular}
\caption{Percentage of KG individuals that are in line with respective CQ answers for different combinations of prompt and CQ answers. (x is the instance with no meaningful KG for that particular combination)}
\label{tab:KGevalresult}
\end{table}

\end{comment}
\begin{table}[htbp]
\centering
\begin{tabular}{| p{4.5cm}| p{1.5cm} | p{1.5cm} | p{1.5cm} | p{1.5cm} |} 
 \hline
 Publication / Combination of prompt and CQ answers & Prompt v1 and CQ answer v1 & Prompt v1 and CQ answer v2 & Prompt v2 and CQ answer v1 & Prompt v2 and CQ answer v2 \\ 
 \hline\hline
 Klein et al. 2015 \cite{klein2015deep} & 24.32 & x & x & 9.42  \\ 
 \hline
 Khalighifar et al. 2021 \cite{Khalighifar2021-ad}  & 85.71 & x & x & 68.06  \\
 \hline
 Choe et al. 2021 \cite{Choe2021-ar} & 73.21 & 64.41 & 59.26 & 64.29 \\
 \hline
 Mahmood et al. 2016 \cite{Mahmood2016-nx} & 91.53 & 81.48 & 82.76 & 77.55 \\
 \hline
 Younis et al. 2020 \cite{Younis2020-ce} & 66.67 & x & 67.51 & 61.67  \\ [1ex] 
 \hline
\end{tabular}
\caption{Percentage of KG individuals that are in line with respective CQ answers for different combinations of prompt and CQ answers. (x is the instance with no meaningful KG for that particular combination)}
\label{tab:KGevalresult}
\end{table}
\vspace{-1cm}
%\sheeba{I have added the csv files for each version in seperate folders in git.I think v10 is more reasonable. Thanks for addressing the comments that I added here though it was meant for me to complete. I will look into manuscript later and refine where required.}
\section{Discussion}
\label{sec:evaluation}
Despite well-known problems such as hallucination, lack of critical thinking, outdated retrieval, and prompt sensitiveness \cite{lu2022fantastically,pan2023large,razniewski2021language}, LLMs are increasingly being used for information retrieval. Initially, we tried to create a KG representing the method information of the DL pipeline from scholarly publications in a single step by providing the whole publication content by prompting the LLM. However, this approach provided little to no content related to KG. Then, we divided the whole process into six components, as discussed in Section \ref{sec:methods}.

The LLM-generated output produced in our pipeline contained excessive unnecessary explanation, which was refined using automatic text processing techniques, as demonstrated in the CQ Answering step, where redundant and repetitive segments of the generated content were eliminated. This helps to minimize the disagreements to 42 out of 200 between the human evaluator and LLM Judge. However, in most instances of disagreement, Judge LLM deemed the human-annotated `partial' labelled CQ answers `wrong'. Furthermore, there are instances where the generated KG format needs adjustments to ensure compatibility with libraries or SPARQL engines for querying purposes.
%\sheeba{Rephrased the below sentence to above paragraph}
%In our (semi-) automated pipeline, we try to automate as much as possible, but human-in-the-loop is much more recommended to mitigate the unnecessary generated content. 
\begin{comment}
    I don't quite understand this: Is this something where you suggest more human in the loop or the argument that the way you do it is good already?
\end{comment}

\begin{comment}
    the paragraph below is difficult to follow
\end{comment}

\begin{comment}
Although our proposed semi-automated pipeline shows the potential for automatic KG generation from the provided content, the evaluation of the generated KG reveals multiple imperfections 1) creating a KG with repetitive individual labels even though the provided CQ answers do not comprise multiple entities, which leads to the creation of no meaningful KG 2) Sometimes the LLM is struggling to capture the ambiguous information from provided content leading to incomplete KG creation. 3) The consistency of the generated KG is not the same because LLM is attributing the rdfs label as "Not Specified" for some KG individuals if it cannot find related information in the provided content, but sometimes it simply omits the KG individuals with no related information in the provided content. 4) Prompt sensitiveness is another drawback while creating KG.
\end{comment}
%\sheeba{Rephrased the above paragraph to include challenges/limitations}
Although our semi-automated pipeline demonstrates the potential for automatic KG generation from provided content, the evaluation of the generated KG reveals a few limitations:
(1) The consistency of the generated KG varies; the LLM assigns the rdfs:label for some KG individuals as `Not Specified' if it cannot find related information in the provided content. In other instances, it simply omits KG individuals with no associated information.
(2) Multiple KG individuals are created for the same ontology class by repeating the same value for the class due to LLM's struggle to capture the intact information from the provided content. In those cases, LLM fails to create meaningful KGs.

Prompt engineering has been pivotal at every stage of our pipeline, particularly in generating RAG CQ answers and KG generation. During KG construction, the outcome varied even with the change in the prompt input order of CQ answers and ontology. %proclaim meaning 
Throughout the entire pipeline, minor alterations in the prompt resulted in the LLM generating different answers most of the time, reflecting a well-known LLM issue that also persists in our pipeline. To address this challenge, we incorporated in-context examples wherever required, which helped to reduce the variability in the generated and hallucinated content. 
Initially, we describe the task briefly for writing an initial prompt. After reviewing the initial results, we added more details to the prompt via trial and error method to reach the optimal prompt, which in turn provided optimal expected answers. 
\begin{comment}
In addition to the Mixtral 8x7B model, we also experimented with other open-source models, Llama 2-70B and Falcon-40B. When manually comparing the quality of the content generated by all these models, Mixtral 8x7B outputs were superior comparatively and provided the expected results. \sheeba{This is added in Methods to show that why we chose Mixtral model, so we should comment it out.}
\end{comment}

We investigated different tools like Ragas\footnote{\url{https://docs.ragas.io/en/latest/index.html}} and Tonic ai\footnote{\url{https://www.tonic.ai/validate}} to evaluate the RAG-generated content; however, they function fully only when an OpenAI API key is provided. Therefore, we devised a judge LLM with open-source LLMs that can rate the generated content using ground truth values.

\section{Conclusion} \label{sec:conclusion}
In this study, we have explored the use of open-source LLMs for the creation of ontologies and knowledge graphs. Our proposed pipeline follows established ontology engineering practices and shows the potential of LLMs acting as assistants (or co-pilots) to human experts. With this, ontology and KG creation require significantly lower effort and less semantic web expertise, making these powerful tools available for broader use.
In our future work, on the one hand, we aim to take advantage of this in our own ontology and KG development work. On the other hand, we will continue improving the pipeline. 
To do so, we plan to run our pipeline on different hardware and evaluate the results using different open-source LLMs to discern potential variations in results.  
The prompt sensitivity of the LLM is high implying that even minor alterations in prompt design could significantly impact LLM outputs. 
Similarly, we anticipate varying results based on hardware usage due to differences in variable initializations across distinct hardware configurations, which can also impact LLM outputs.
Furthermore, we will explore methods for mapping the generated ontology with other ML/DL ontologies.
\bibliographystyle{splncs04}
\bibliography{main}

\end{document}